\documentclass[runningheads]{llncs}

\usepackage{graphicx}
\usepackage{graphicx,url}
\usepackage{colortbl}
\usepackage[utf8]{inputenc}
\usepackage{verbatim}
\usepackage{amsmath,amssymb,amsfonts}
\usepackage{algorithmic}
\usepackage{graphicx}
\usepackage{textcomp}
\usepackage{xcolor}
\usepackage{cleveref}
\usepackage{makecell}
\usepackage{multirow}
\usepackage[flushleft]{threeparttable}
\usepackage{hhline}

\newcolumntype{P}[1]{>{\centering\arraybackslash}p{#1}}

\usepackage{todonotes}

\title{Portuguese Man-of-War Image Classification with Convolutional Neural Networks}
\author{Alessandra Carneiro\inst{1}, Lorena Nascimento\inst{2}\\ Mauricio Noernberg\inst{2}, Carmem Hara\inst{1}, Aurora Pozo\inst{1}}
\institute{Departamento de Ciência da Computação\inst{1} and Centro de Estudos do Mar \inst{2}\\ Universidade Federal do Paraná,  Curitiba -- PR -- Brazil \\
\email{ \{alessandra.k,lorena.sn,m.noernberg\}@ufpr.br, \{carmem,aurora\}@inf.ufpr.br}}

\begin{document}
\maketitle

\begin{abstract}
Portuguese man-of-war (PMW) is a gelatinous organism with long tentacles capable of causing severe burns,  thus leading to negative impacts on human activities, such as tourism and fishing. There is a lack of information about the spatio-temporal dynamics of this species. Therefore,  the use of alternative methods for collecting data can contribute to their monitoring.  Given the widespread use of social networks, data present on the social media Instagram could represent a promising methodology for monitoring such species. This research  aims to validate the use of convolutional neural networks for the Portuguese man-of-war images classification allowing the automatation of  the recognition of the Istagram records that contain this species. We created  a suitable dataset, and we trained three different neural networks: VGG-16, ResNet50, and InceptionV3, with and without a pre-trained step with the ImageNet dataset. We analyzed the results of the  different neural networks   through the accuracy, precision, recall, and F1 score metrics. The pre-trained ResNet50  network presented the best results , obtaining 94\% in the accuracy metric and 95\% in the precision, recall, and F1 score metrics. Therefore, we point out that convolutional neural networks can be very effective for recognizing  Portuguese man-of-war  images  from the social media Instagram.
\end{abstract}

\section{Introduction}

Jellyfish stings are coastal hazards being some of the most common envenomation encountered by humans in marine ecosystems. Among the most dangerous species, the jellyfish Portuguese man-of-war (PMW) (\textit{Physalia physalis}) represents a public health issue, a threat to tourism in many parts of the world \cite{labadie2012portuguese}. Their stings can cause intense immediate skin pain and elicit serious systemic effects, including headache, vomiting, abdominal pain, diarrhea, muscular, cardiac, neurological, and allergic reactions, causing even deaths occasionally \cite{burnett1989fatal}.

Portuguese man-of-war has a gas-filled float that keeps it floating passively at the water surface with tentacles that can be more than 30 meters long containing the venom trailing below. Stings events are usually associated with wind conditions, local currents, and the consequent transport of these animals toward the coast. The abundance and occurrence of this species fluctuate inter-annually, but the highest risk of stings occurs during the warmer months, due to the peak of bathers in coastal environments during the holiday season \cite{haddad2013environmental}.

Information about Portuguese man-of-war occurrence and distribution is essential to implement measures of prevention and manage stings risks, such as beach closures, warning flags, anti-jellyfish nets \cite{piraino2016antijellyfish}, and forecasting systems \cite{macias2021model}. However, the Portuguese man-of-war is one of the least studied jellyfish. Gaps are caused partially due to the difficulty of studying this species, hampering its capture in traditional methodologies \cite{headlam2020insights}.

 Scientists are increasingly utilizing information provided by citizens \cite{marambio2021unfolding}  to address knowledge gaps and acquire occurrence records of this species.  \cite{nascimento2020monitoring} analyzed the feasibility of using data extracted from social media to complement the information on the distribution of Portuguese man-of-war on the Brazilian coast. 
For this, a methodology for collecting, filtering, and storing Instagram data was established. From specific hashtag searches, evidence such as photos and videos of the animal were sought in posts, as well as, information about the location and date of occurrence. However, the manual methodology of social media data filtering suffers the consequences of human limitations, resulting in a tedious and time-consuming process. Furthermore, the continuity and expansion of the method require a permanent human effort, making it a little scalable.
Computational techniques, such as machine learning and artificial intelligence, are improving the ability to search, collect and process crowdsourcing ‘big data’ from social media in a less time-consuming way, also  helping to create a centralized open-source data system \cite{vanoast2017animal}.

The media treatment consists of verifying the presence of a Portuguese man-of-war in at least one of the images or videos of the social media post. Therefore, we propose to model the problem as a binary classification of images between a positive class for the presence of  jellyfish, and a negative class represented by any other random image. This task is complex, and a suitable approach would be the use of Convolutional Neural Networks (CNN), a deep neural network that showed good previous results in complex image classification \cite{russakovsky2014imagenet}. In the present study, we aim to implement and analyze some models of CNN applied to the image classification problem of the Portuguese man-of-war, to validate the use of this technology for the methodology defined by \cite{nascimento2020monitoring}.

This paper is organized as follows. Section \ref{sec:cnn} covers the technical approach used  for image classification. Related work is described in Section \ref{sec:relatedwork}. Then, the methodology used for the development is described in Section \ref{sec:matherialsandmethods}. Section \ref{sec:results} presents and discusses the results obtained from the experiments. Finally, Section \ref{sec:conclusion} concludes the paper with a discussion of the results and future directions.

\section{Convolutional Neural Networks}\label{sec:cnn}

\begin{figure}[!htb]
       \centering
        \includegraphics[width=\textwidth /2]{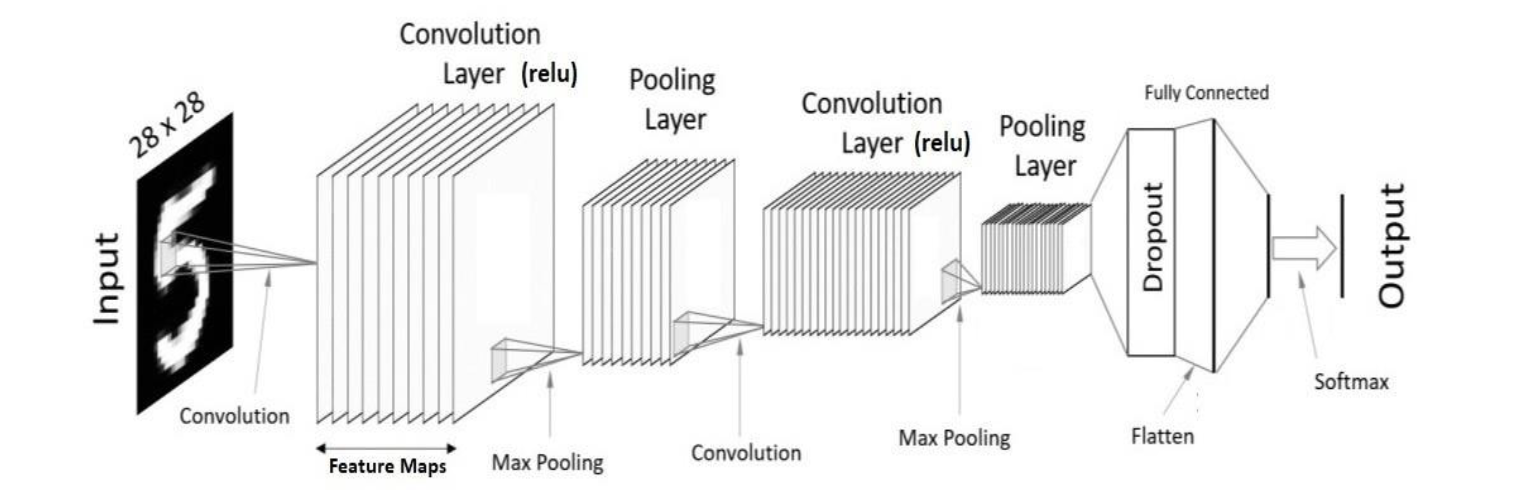}
        \caption{Example of a convolutional neural network architecture.\cite{shyam2021convolutional}}
        \label{fig:cnn}
\end{figure}

There are several deep neural network architectures with a wide variety of applications. According to \cite{alom2019state}, one of the architectures with the most promising results in the area of image classification is the Convolutional Neural Network (CNN). CNNs gained popularity in 2012 for their use in the popular ImageNet Large Scale Visual Recognition Challenge (ILSVRC) object recognition competition \cite{russakovsky2014imagenet}, where they achieved very promising results compared to other technologies. One of the main factors for the success of CNNs is their ability to automatically extract features from the provided images, unlike other approaches that usually require manual feature extraction, such as support vector machines (SVM) \cite{alom2019state}. The architecture of a CNN can have several formats, depending on its implementation. As a standard, the input of a CNN always goes through several processes sequentially, and each step of these processes is called a layer. The layers can be convolutional, pooling, normalization, fully-connected, loss, and other types \cite{alom2019state}. A generic architecture of a CNN is shown in Figure \ref{fig:cnn}.
The convolutional layers, which give the network its name, are meant to extract features from its input and form a feature map. Each input is convoluted with filters that can learn and passes through an activation function, generating an output feature map, which serves as input to the next layer\cite{alom2019state}.

CNNs, in general, require large image databases, such as ImageNet, to achieve good results \cite{alom2019state}. So, transfer learning has become a popular technique. In transfer learning, a neural network is trained on problem $P1$, where there is a large image database. After then,  this network is used to solve $P2$.  This technique can achieve good results if both problems have related inputs. Moreover, it can help to improve the learning of useful features and generalization of the second problem \cite{goodfellow2016deep}. Finally, a fine-tuning is applied, where adjustments are made to the pre-trained network to better adapt to the new problem. These adjustments usually include replacing the original output layer with an output layer that fits the new problem, as well as inserting new fully connected layers and regularization techniques. In addition to making changes to the network structure, a total or partial freezing of the pre-existing layers is performed. This procedure aims to prevent the network from having its old weights modified with the new training data, and consequently losing its existing robust learning.

Currently, there are several popular CNN architectures considered to be state-of-the-art, as they can achieve good performance on different classification tasks \cite{alom2019state}. Performance comparison of architectures applied on large image databases has been fundamental in guiding the development of models applied on varied tasks \cite{alom2019state}. A widely used competition in this regard is ILSVRC \cite{russakovsky2014imagenet}, which uses the ImageNet image database consisting of over 14 million images. Some architectures described here have gained prominence in this competition. The VGG (Visual Geometry Group) architecture defined in \cite{simonyan2014deep} is a trained model that was highlighted in the competition in 2014. It has six types of layers and a total of 13 convolutional layers and 3 fully-connected layers. Its variation, called VGG-16 and VGG-19, consists of two models that differ by the number of convolutional layers. The ResNet architecture was the winner of the ILSVRC challenge in 2015 in the categories of image classification, detection, and localization, in addition to other recognized challenges \cite{he2016deep}. It can be created with various numbers of layers, but some of its most popular patterns are RestNet50, with 50 layers, and ResNet101, with 101 layers. Defined in \cite{szegedy2015deeper}, the Inception architecture was created to optimize the use of computational resources, so that the network can become more complex without becoming so computationally costly. The architecture has some variations, such as InceptionV1, InceptionV2, and InceptionV3.

\section{Related work}\label{sec:relatedwork}
\begin{figure}[!htb]
       \centering
        \includegraphics[width=\textwidth /2]{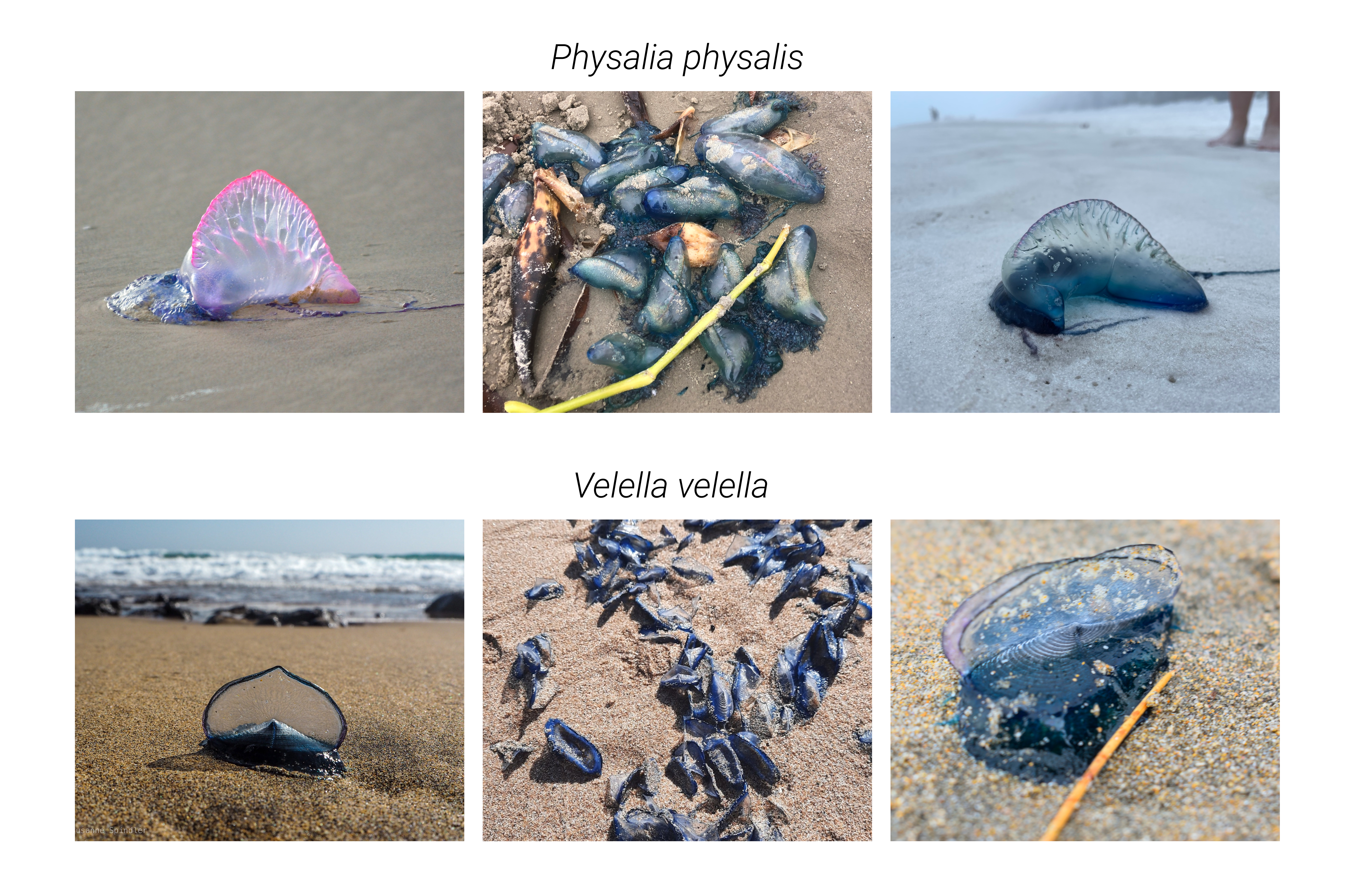}
        \caption{Comparative images of \textit{Physalia physalis} and \textit{Vellela vellela}.}
        \label{fig:PhysaliaAndVellela}
\end{figure}

This section will present the main related works that underpin the present work. First, the research related to Portuguese Man-of-war will be presented, focusing on their visual characteristics. Next, the researchers found focusing on deep learning applications of jellies will be described.
For the task of image classification, it is important to perform an analysis of the main physical characteristics of the object of study and in which environments it is commonly photographed, to understand which images compose the database, and to identify the possible errors that can be made by the neural networks. The Portuguese Man-of-war, \textit{Physalia physalis}, are organisms that live in a colony and belong to the cnidarian group, of the order siphonophores. The individuals that make up the colony are dependent on each other, and each has a specific function \cite{munro2019morphology}. The greatest visual highlight of its composition is its pneumatophore, a balloon-shaped structure filled with the gas used in navigation, as it has vivid colors in pinkish and bluish tones \cite{bardi2007taxonomic}. Despite being quite characteristic, the colors and the shape of the caravela-portuguesa can be confused with another cnidarian: the \textit{Velella velella}, popularly known as By-the-wind Sailor. Both species are oceanic and float freely on the sea surface, and eventually are carried to shore and stranded on the sand \cite{badalamenti2021portuguese}. Analyzing Figure \ref{fig:PhysaliaAndVellela}, it is possible to observe that the images of the Portuguese Man-of-war have a more bluish coloration and are in groups that can be easily confused with the images of the By-the-wind Sailor, even by humans.
Few studies were found that specifically focused on the classification of Portuguese Man-of-war images, therefore, we describe studies of any species of jellyfish or gelatinous organism, animals physically similar to the Portuguese Man-of-war. In \cite{martinabadal2020jellytoring} the Jellytoring system is presented, which aims to detect and quantify different jellyfish species through a CNN, allowing the generation of a history of the presence of these animals. The input images for this model come from underwater video recordings, and refer to three species: \textit{Pelagia noctiluca}, \textit{Cotylorhiza tuberculata}, and \textit{Rhizostoma pulmo}. The architecture used was Inception-Resnet-v2 pre-trained with the COCO database, which was compared with the ResNet101 and InceptionV2 architectures, and achieved F1 Score metrics greater than 95\%. The study presented in \cite{gao2021realtime} analyzes the problem of detecting and classifying seven species of jellyfish and fish in underwater images, to promote the monitoring of these species. The architecture employed was YOLOv3, a CNN used to detect objects in an image. The authors make a series of improvements in the application of the CNN and manage to achieve the result of the accuracy metric at 95.53\%.
\cite{mcliwaine2021jellynet} presents JellyNet, a CNN model trained to detect jellyfish blooms from high-resolution images obtained by unmanned aerial vehicles (UAV), also known as drones. The architecture used was based on the VGG-16, and achieved 90\% accuracy.

In \cite{record2018jelly}, a system for predicting the occurrence of three jellyfish species, \textit{Cyanea capillata}, \textit{Aurelia aurita}, and \textit{Staurostoma mertensii}, in the Gulf of Maine is tested. The system used data from satellites and also data submitted by citizens via email and the social network Twitter to train a prediction model. To engage citizens to collect and submit data for the research, announcements were made through social media and institutional networks asking for submissions. In total, 259 records were collected via email and 29 via Twitter. Overall, the model showed promising results. However, the authors point out that these results do not necessarily imply good actual forecasting performance, and that the generality of the model will become clearer as forecasts over longer periods are included. The study presented in \cite{correia2020automatic} plans to develop a low-cost system for the detection, recognition, and classification of various types of incidents in coastal areas through UAVs. The incidents include everything from oil spills to the appearance of dangerous sea creatures, such as the Portuguese Man-of-war. The first part of the solution is composed of a CNN to process the images obtained by the UAVs, and the second part is a mobile application to manage the UAVs. The solution is still being developed by the authors.

In sum, despite the similarity between the present work and the cited works, due to the physical characteristics of Portuguese Man-of-war being similar to that of other jellyfish, almost none of the models could be applied in the classification problem presented here, because they have another function. In \cite{martinabadal2020jellytoring} and \cite{mcliwaine2021jellynet}, the goal is to detect jellyfish, i.e., the intention is to find where and how many animals are in the image. In \cite{gao2021realtime}, the goal is also to detect jellyfish and classify them, but using underwater images and among different species than the one studied here. In \cite{record2018jelly}, the function of the model is to predict the appearance of jellyfish in a region. In \cite{correia2020automatic}, the model has a broader objective than the one presented here, but still uses images of Portuguese Man-of-war. However, it is not possible to validate whether the model can serve as a basis for the problem of this study as it is still under development. Furthermore, the use of underwater images or aerial images for training is a very relevant factor, since most of the images of jellyfish obtained from Instagram are of the animals stranded on the sand, making them quite distinct. Moreover, data collection, in most of the works cited, was performed in an active manner, where a job is employed to perform this task. In the present work, the task is to analyze a large amount of data present on social media, specifically Instagram. Such data is not collected specifically for this purpose, but is a useful source of information about biodiversity, and is cheap and easy to collect.

\section{Materials and methods}\label{sec:matherialsandmethods}

This section describes the methods used for image collection and implementation of the neural networks used in this study.
\subsection{Data Collection}
The database was grouped into two classes: the Portuguese Man-of-war (PMW) class, composed of images that contain some appearance of the Portuguese Man-of-war, and the not-Portuguese Man-of-war (not-PMW) class, composed of images that do not contain any appearance of the Portuguese Man-of-war. The PMW class was based mainly on images of Portuguese Man-of-war stranded on the sand, and the not-PMW class was created based on seven types of images: person (mostly photographed on beaches), ships (mostly of the caravel type), illustrations, tattoos, \textit{Velella velella}, jellyfish of various species, and random images. 

To collect the data from the PMW class, two data sources were used. The first source was the data obtained from Instagram by \cite{nascimento2020monitoring}, through the search for the hashtag \#caravelaportuguesa, where 426 images were collected. The second source was the iNaturalist platform, which allows people around the world to register and share their records of various species of animals and plants. The platform allows the data to be exported using filters, such as species and data type, which allowed 5,731 images of Portuguese Man-of-war to be obtained easily and efficiently. In total, 6,157 images were obtained for the PMW class.

For the not-PMW class data collection, three data sources were used. The first was the iNaturalist platform, where 2,876 images of \textit{Velella velella} and 649 images of random jellyfish (not including the Portuguese Man-of-war)  recorded in South America. The second data source was the Bing platform, a search engine that allows the return of images by searching for a term. To collect the largest number of images in the shortest time, a script written in Python language was used to search the Bing platform and save the resulting images. The number of images and the searched terms can be seen in Table \ref{table:imagesBing}. 
\begin{table}[]
\centering
\caption{Images taken by Bing.}
\label{table:imagesBing}
\begin{tabular}{|l|l|l|}
\hline
\textbf{Image type} & \textbf{\begin{tabular}[c]{@{}l@{}}Search terms \\ (translated from Portuguese)\end{tabular}}             & \textbf{\begin{tabular}[c]{@{}l@{}}Image \\ amount\end{tabular}} \\ \hline
Person              & "selfie", "person on the beach"                                                                           & 453                                                              \\ \hline
Ship                & "portuguese caravel ship", "ship"                                                                         & 471                                                              \\ \hline
Illustration        & \begin{tabular}[c]{@{}l@{}}"Portuguese man-of-war illustration", \\ "jellyfish illustration"\end{tabular} & 585                                                              \\ \hline
Tattoo              & \begin{tabular}[c]{@{}l@{}}"Portuguese man-of-war tattoo", \\ "tattoo"\end{tabular}                       & 697                                                              \\ \hline
Random              & "animals", "handicraft", "beach"                                                                           & 602                                                              \\ \hline
\end{tabular}
\end{table}
The last data source used was the social media Instagram itself, from the results obtained by searching the hashtag \#caravela (caravel in portuguese). This term was used because it has results very related to the hashtag \#caravelaportuguesa, but mostly composed of images that are not a record of \textit{Physalia physalis}: ships, illustrations, tattoos, people and random. This method obtained a total of 2,329 valid images for the class not-PMW. To keep the base as similar as possible to the negative results obtained for the search for \#caravelaportuguesa, only 500 images of \textit{Velella velella} and 500 images of random jellyfish were selected to compose the final database, which was added with images obtained through the Bing and Instagram platforms. A  total 6,137 images were collected.

All images went through pre-processing and were resized to a size of 224x224, to reduce the computational cost. This size was chosen because it is the standard input for the VGG-16 and ResNet50 architectures. The default size of the InceptionV3 architecture is 299x299, which causes the size of the feature maps generated by the network to change.
The collection of images of the not-PMW class obtained on sources such as the Instagram and Bing platforms does not have a secure trust to ensure that the images belong to the desired categories. To maintain the integrity of the class, a manual cleanup was performed, removing images that represented a Portuguese man-of-war. However, like any manual task, this is a flawed process, and some invalid images may have been kept as outliers in the base.

\subsection{Model training}

The CNNs selected for evaluation were VGG-16, ResNet50 and InceptionV3, through the metrics of accuracy, precision, recall and F1 Score. The selected neural networks were implemented in Python language, widely used in solving machine learning problems due to its large number of libraries in the area. The code was written using the Jupyter Notebook tool, to facilitate visualization of the results and generation of the metrics. The main tool used was the Keras API, belonging to the Tensorflow library, which provides the implementation of several popular CNN models with or without pre-training. 

The training of the networks was run on an Apple M1 processor machine with an 8-core CPU and 8-core GPU and 16GB of RAM. Each architecture was run 2 times, the first time without any pre-training, and the second time using ImageNet database pre-training. The evaluation of the results obtained was done using the metrics accuracy, precision, recall and F1 Score.

\begin{figure}[!htb]
       \centering
        \includegraphics[width=\textwidth /2]{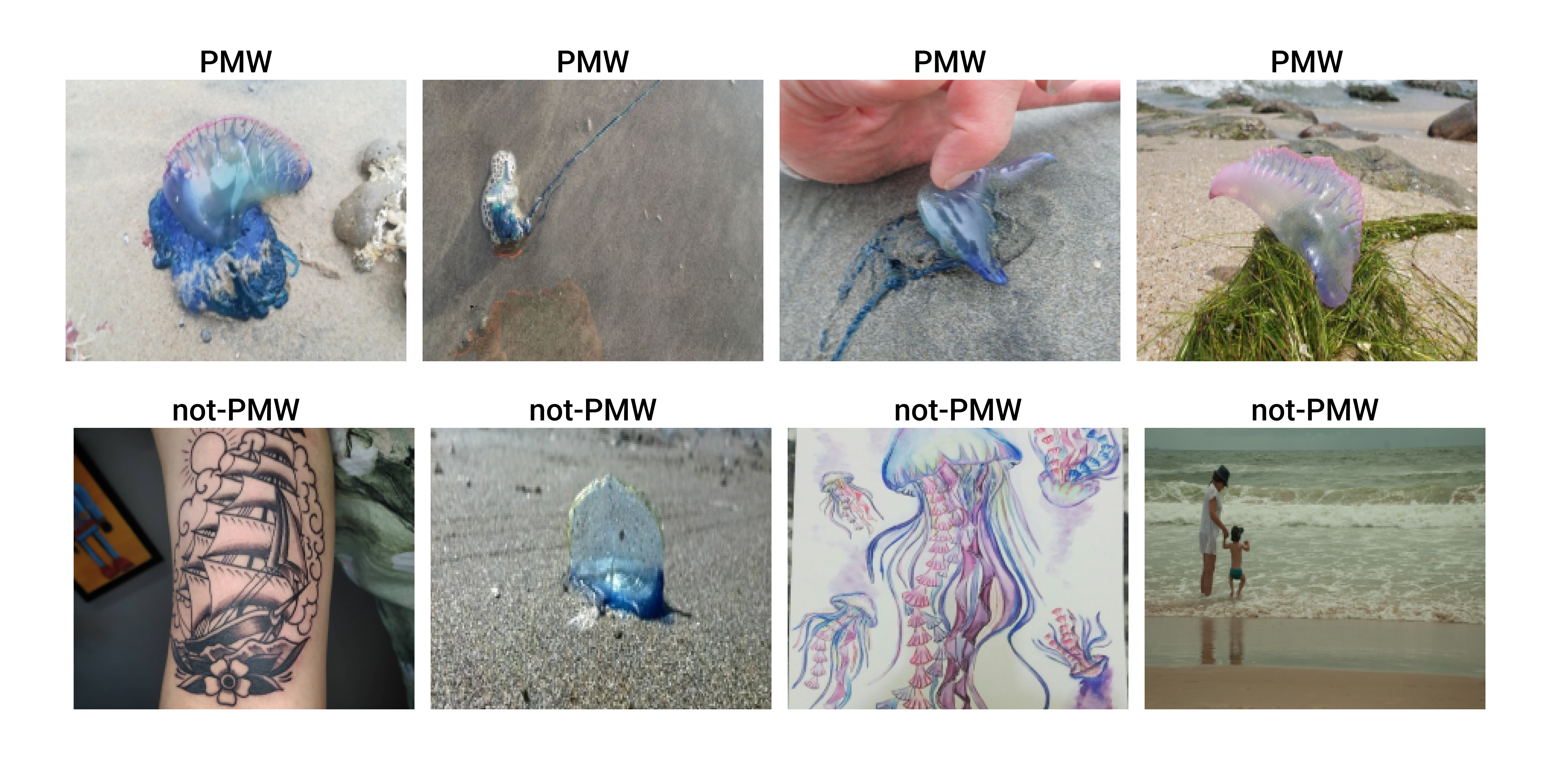}
        \caption{Sample images present in the training database}
        \label{fig:trainingSample}
\end{figure}

The database used in the experiments was divided into training, containing about 60\% of the total database, validation, with 20\%, and testing, also with 20\%. Each plot was manually split so that the not-PMW class remained relatively balanced among its image types. A sample of the images present in the training database can be seen in Figure \ref{fig:trainingSample}.
We used the data augmentation technique to increase the number of images presented in the training base, applying rotation, zoom, and horizontal mirroring randomly on the images. The same fine-tuning approach was used in all tested CNNs, and therefore, all had the same modifications. One of the main changes was the replacement of the output layer with a new output layer suitable for the proposed problem, composed of an average pooling layer, a flattening layer, and a fully-connected layer that uses the ReLu activation function, and a dropout regularization layer with a percentage of 30\%. In addition, all pre-existing layers in the CNNs were frozen during the training of the networks using ImageNet pre-training. The loss function used was the cross-entropy, more specifically, the binary cross-entropy, a function provided by Keras suitable for binary classification problems. The training was performed with a threshold of 50 epochs, coupled with an Early Stopping based on the loss obtained at the validation. 

\section{Results and Discussion}\label{sec:results}

\begin{figure}[!htb]
       \centering
        \includegraphics[width=\textwidth /2]{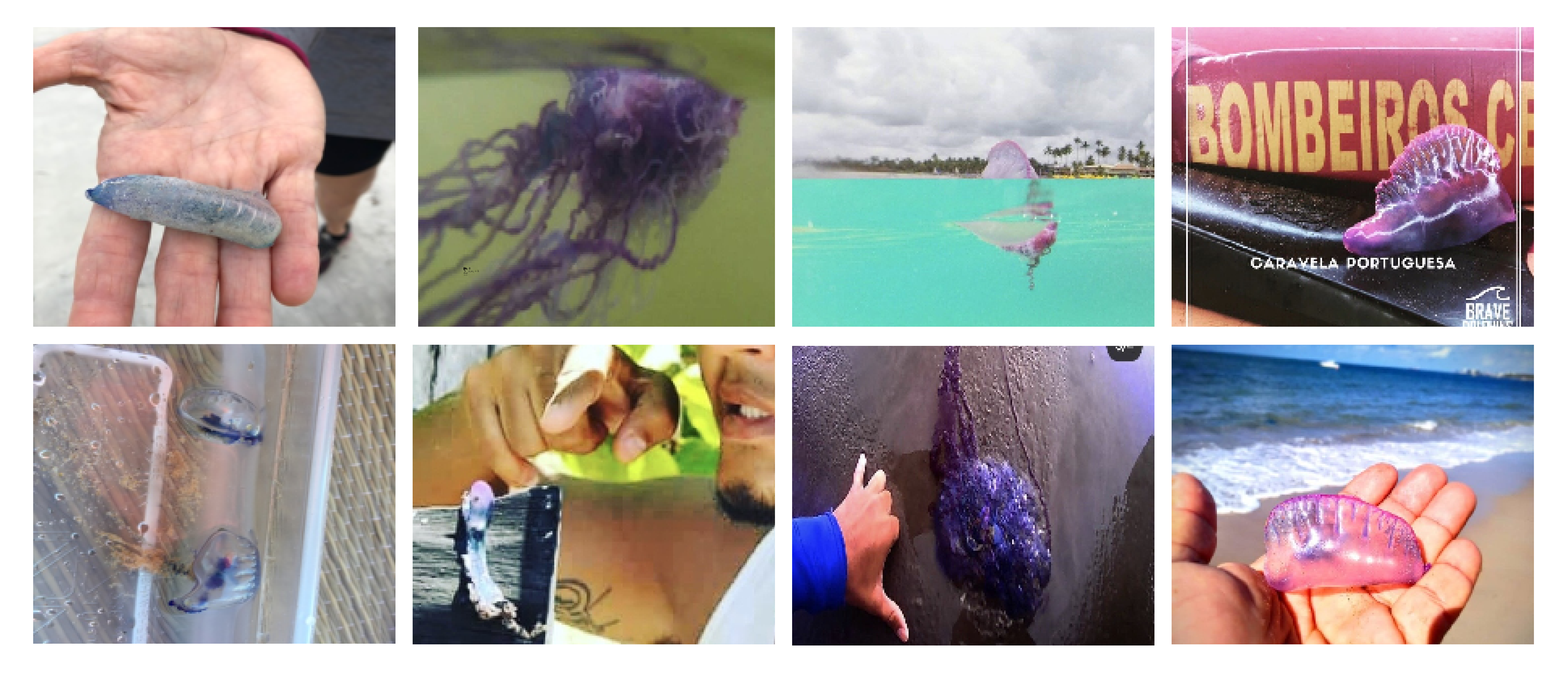}
        \caption{PMW class false negatives sample.}
        \label{fig:falseNegatives}
\end{figure}

\begin{figure}[!htb]
       \centering
        \includegraphics[width=\textwidth /2]{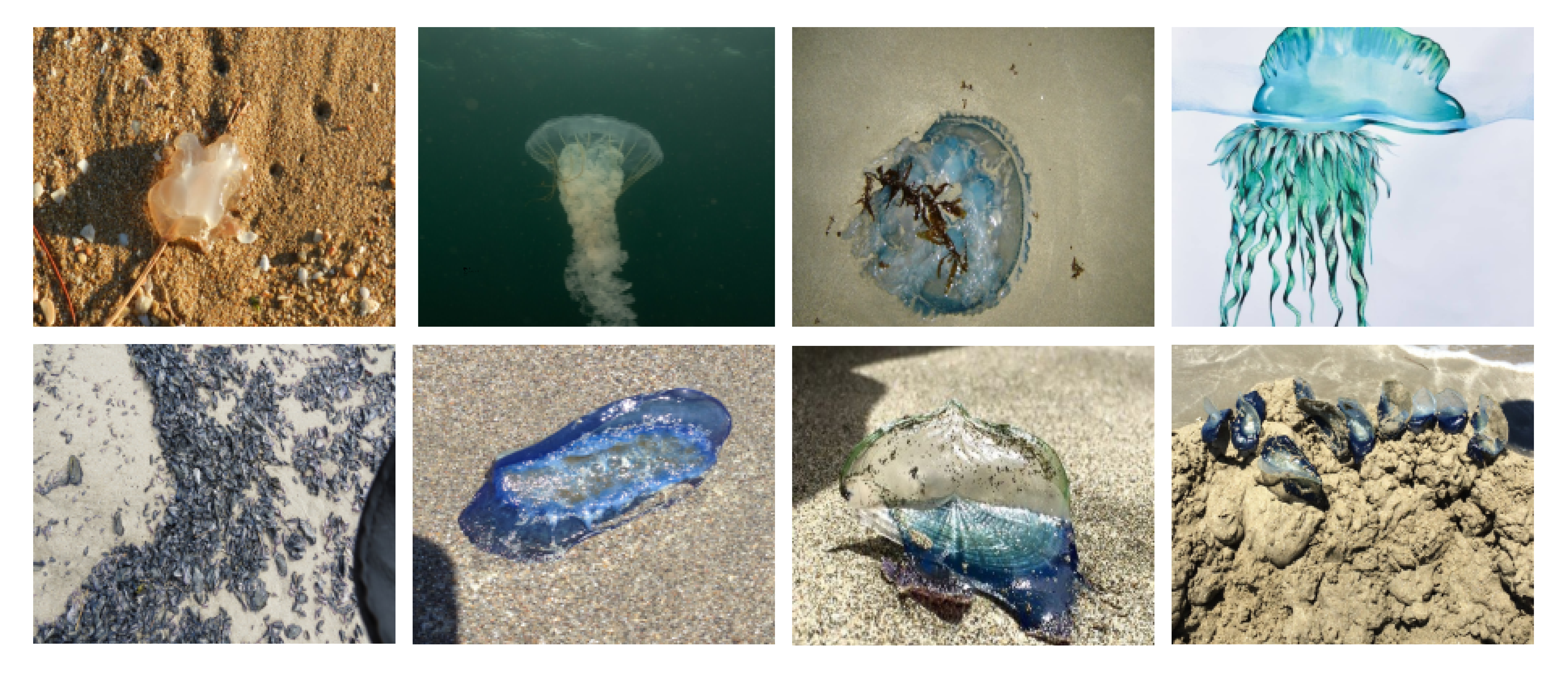}
        \caption{PMW class false positives sample.}
        \label{fig:falsePositives}
\end{figure}

The results obtained by the different CNNs are presented in the following tables. The tables have the same format: the first three rows represent the architectures without pre-training, and the last three rows represent the architectures pre-trained with the ImageNet database.
\begin{table}
\centering
\caption{Average test results for each CNN.}
\label{table:avgtestresults}
\begin{tabular}{|l|l|l|l|l|} 
\hline
\textbf{Architecture}                                 & \textbf{Accuracy} & \textbf{Precision} & \textbf{Recall} & \textbf{F1 Score}  \\ 
\hline
VGG-16                                                & 0.7482            & 0.75               & 0.75            & 0.75               \\ 
\hline
ResNet50                                              & 0.8360            & 0.84               & 0.84            & 0.84               \\ 
\hline
InceptionV3                                           & 0.8238            & 0.82               & 0.82            & 0.82               \\ 
\hline
VGG-16 + ImageNet                                     & 0.9308            & 0.93               & 0.93            & 0.93               \\ 
\hline
\rowcolor[rgb]{0.902,0.902,0.902} ResNet50 + ImageNet & 0.9475            & 0.95               & 0.95            & 0.95               \\ 
\hline
InceptionV3 + ImageNet                                & 0.9386            & 0.94               & 0.94            & 0.94               \\
\hline
\end{tabular}
\end{table}
\begin{table}
\centering
\caption{Results per class obtained in the tests of each CNN.}
\label{table:testresultsperclass}
\arrayrulecolor{black}
\begin{tabular}{|l|l|l|l|l|l|} 
\hline
\textbf{Architecture}                                                                                                & \textbf{Class} & \textbf{Accuracy} & \textbf{Precision} & \textbf{Recall} & \textbf{F1 Score}  \\ 
\hline
\multirow{2}{*}{\textbf{VGG-16}}                                                                                     & PMW            & 0.78              & 0.73               & 0.79            & 0.76               \\ 
\cline{2-6}
                                                                                                                     & not-PMW        & 0.70              & 0.77               & 0.71            & 0.74               \\ 
\hline
\multirow{2}{*}{\textbf{ResNet50}}                                                                                   & PMW            & 0.89              & 0.80               & 0.89            & 0.84               \\ 
\cline{2-6}
                                                                                                                     & not-PMW        & 0.77              & 0.88               & 0.78            & 0.83               \\ 
\hline
\multirow{2}{*}{\textbf{InceptionV3}}                                                                                & PMW            & 0.82              & 0.83               & 0.82            & 0.82               \\ 
\cline{2-6}
                                                                                                                     & not-PMW        & 0.82              & 0.82               & 0.83            & 0.82               \\ 
\hline
\multirow{2}{*}{\textbf{VGG-16 + ImageNet}}                                                                          & PMW            & 0.93              & 0.93               & 0.93            & 0.93               \\ 
\cline{2-6}
                                                                                                                     & not-PMW        & 0.93              & 0.93               & 0.93            & 0.93               \\ 
\hline
\rowcolor[rgb]{0.902,0.902,0.902} {\cellcolor[rgb]{0.902,0.902,0.902}}                                               & PMW            & 0.95              & 0.94               & 0.96            & 0.95               \\ 
\hhline{|>{\arrayrulecolor[rgb]{0.902,0.902,0.902}}->{\arrayrulecolor{black}}-----|}
\rowcolor[rgb]{0.902,0.902,0.902} \multirow{-2}{*}{{\cellcolor[rgb]{0.902,0.902,0.902}}\textbf{ResNet50 + ImageNet}} & not-PMW        & 0.93              & 0.96               & 0.94            & 0.95               \\ 
\hline
\multirow{2}{*}{\textbf{InceptionV3 + ImageNet}}                                                                     & PMW            & 0.95              & 0.93               & 0.95            & 0.94               \\ 
\cline{2-6}
                                                                                                                     & not-PMW        & 0.92              & 0.95               & 0.93            & 0.94               \\
\hline
\end{tabular}
\end{table}
Analyzing the results presented in  Table \ref{table:avgtestresults} and Table \ref{table:testresultsperclass}, it is possible to observe that the networks pre-trained with the ImageNet database obtained considerably better results than networks without any pre-training. 
Among the networks without any pre-training, the best performance was obtained by the ResNet50 network, with about 83\% accuracy, followed by InceptionV3, with 82\%. The VGG-16 network had the worst performance, with results well below the other networks, with only 74\% accuracy. 
Among the pre-trained networks, all obtained very consistent metrics in both classes, demonstrating a generalized quality of the models. Although very close, the best performance was obtained by the ResNet50 network, with almost 95\% of accuracy Therefore, its results will be analyzed in more depth below.
The pre-trained ResNet50 obtained accuracy in the PMW class slightly better, with an error rate of about 4.30\%, while the not-PMW class obtained an error rate of about 6.18\%, i.e., almost 2\% different from the previous one.  There are 53 false negatives in the PMW class between them, 11 have a person in the photo, and 14 have the presence of a human hand, either holding the Portuguese Man-of-war or not. The presence of a human may be a factor impacting the classification since the not-PMW class is also composed of images of people. Besides, the author has difficulty classifying 12 of these images, because they are pictures taken far away from the animal, with low quality, or only of its tentacles.  Figure \ref{fig:falseNegatives} shows an example of a false negative. 
There are 76 false positives, 46 are \textit{Velella velella} images, 25 are jellyfish images, 5 images are illustrations, and one image is of a handicraft. Since the test base is approximately balanced among the image types, the highest error rate for the not-PMW class model is in the \textit{Velella velella} and jellyfish images, which add up to more than 93\% of the total errors. Figure \ref{fig:falsePositives} shows a sample of the false positives.
A possible attempt to improve the performance of the CNN is providing higher examples of images that are part of the groups mentioned with higher occurrence in the false positives and false negatives of the PMW class. However, it is necessary to observe if this change does not cause negative impacts on the classification of other images and consequently worse results.

\section{Conclusion}\label{sec:conclusion}
This study presents the evaluation of the use of convolutional neural networks to automate a step of the methodology proposed by \cite{nascimento2020monitoring}. Three different types of architectures were evaluated to measure their performance in a binary image classification task for the identification of the Portuguese Man-of-war.
The experiments performed show that CNNs present promising results for this task, reaching an accuracy of over 94\% by the best performing network, the ResNet50 pre-trained on the ImageNet database. Therefore, these obtained results show that the application of CNNs is suitable and can be highly beneficial for the collection of evidence of images of Portuguese caravels on Instagram.
Some interesting future directions of this research include the application of different techniques to further improve the results, such as other fine-tuning, data augmentation, and regularization approaches, as well as, the refinement and expansion of the database, focusing on the treatment of the types of image that represented the higher error rate. In addition, other reference architectures can also be evaluated on this problem.
Another proposal for future work is to analyze the automation of other parts of the publication of the Istagram. A complete flow could consist of a system that periodically and automatically collects data, analyzes the data present in each publication, both images and text, and classifies it as a valid or invalid record for the Portuguese Man-of-war. For valid records, the system saves the relevant information in a database. It is also interesting to develop a dashboard that allows  users to manage the data and the system.

\bibliographystyle{splncs04}
\bibliography{references}

\end{document}